\pdfoutput=1

\documentclass[11pt]{article}

\usepackage[preprint]{acl}

\usepackage{times}
\usepackage{latexsym}
\usepackage{epigraph}
\usepackage{amsmath}
\usepackage{todonotes}
\usepackage{listings}
\usepackage{booktabs}
\usepackage{multirow}
\usepackage{multicol}

\usepackage{algorithm}
\usepackage{algpseudocode}
\usepackage{amsmath}
\usepackage{subcaption}

\usepackage{amssymb}

\definecolor{darksnow}{HTML}{335479}
\definecolor{darkersnow}{HTML}{0D2235}
\definecolor{dark}{HTML}{929292}
\definecolor{backcolour}{HTML}{F7F7F7}
\definecolor{backcolour2}{HTML}{F0F8FC}

\definecolor{light}{HTML}{E2E2E2}
\DeclareMathOperator*{\argmax}{arg\,max}

\newcommand{\na}{\textcolor{light}{$\times$}}

\newcommand{\midsepremove}{\aboverulesep = 0mm \belowrulesep = 0mm}
\midsepremove
\newcommand{\midsepdefault}{\aboverulesep = 0.605mm \belowrulesep = 0.984mm}
\midsepdefault
    
\usepackage[T1]{fontenc}

\usepackage[utf8]{inputenc}

\usepackage{microtype}

\usepackage{inconsolata}

\usepackage{graphicx}

\title{Query and Conquer: Execution-Guided SQL Generation}

\author{Łukasz Borchmann \and Marek Wydmuch \\
  Snowflake AI Research \\
  \texttt{lukasz.borchmann@snowflake.com}}

\begin{document}
\maketitle
\begin{abstract}
We propose a novel approach for generating complex outputs that significantly improves accuracy in text-to-SQL tasks. Our method leverages execution results to select the most semantically consistent query from multiple candidates, enabling smaller, cost-effective models to surpass computationally intensive reasoning methods such as o1, o3-mini, and DeepSeek R1 while reducing inference cost by as much as 30 times. 
It integrates effortlessly with existing models, offering a practical and scalable pathway to state-of-the-art SQL generation.
\end{abstract}

\section{Introduction}

Large language models frequently produce correct outputs when multiple samples are considered (\texttt{pass@k}), yet their one-shot accuracy (\texttt{pass@1}) remains significantly lower.
This gap motivates exploring methods that leverage multiple outputs to reliably identify the most promising one.

A prime example is the self-consistency method, which generates a diverse set of reasoning paths from the model and then employs majority voting on the final answers (e.g. numerical values for mathematical problems) to select the most likely correct output \cite{wang2023selfconsistencyimproveschainthought}.
While practical for short, well-defined answers, this strategy quickly breaks down when outputs have multiple correct yet structurally distinct representations, such as in code generation tasks. 

Consider a simple SQL query retrieving all unique department IDs from an `employees' table:

\begin{lstlisting}[language=SQL]
    SELECT DISTINCT department_id
    FROM employees;
\end{lstlisting}

\noindent This query can be restructured in a variety of different yet equivalent forms, including:

\begin{lstlisting}[language=SQL]
    SELECT department_id
    FROM employees
    GROUP BY department_id;
\end{lstlisting}

\noindent Alternatively, it can be expressed in an entirely different way, such as:

\begin{lstlisting}[language=SQL]
    SELECT department_id
    FROM (
      SELECT department_id, ROW_NUMBER() OVER (PARTITION BY department_id ORDER BY department_id) AS rn
      FROM employees
    ) AS subquery
    WHERE rn = 1;
\end{lstlisting}

Each of these variations produces the same result but differs in structure, rendering majority voting strategies ineffective.
We argue that overcoming this challenge demands methods that measure equivalence at the execution level rather than depending on structural comparison and propose such a strategy substantially narrowing the gap between \texttt{pass@1} and \texttt{pass@k} accuracy.

\begin{figure}[tb]
    \centering
    \includegraphics[width=\linewidth]{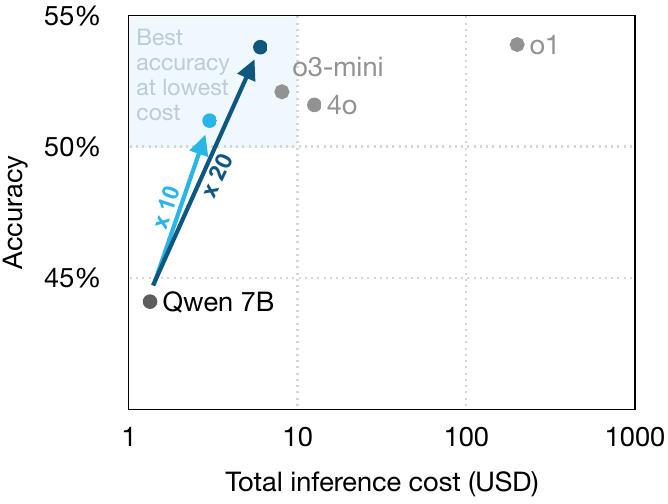}
    \caption{Cost-accuracy analysis for Qwen 2.5 Coder 7B, with or without self-consistency (10-20 samples), compared alongside OpenAI models.}
    \label{fig:cost}
\end{figure}

\begin{figure*}[ht]
    \centering
    \includegraphics[width=\linewidth]{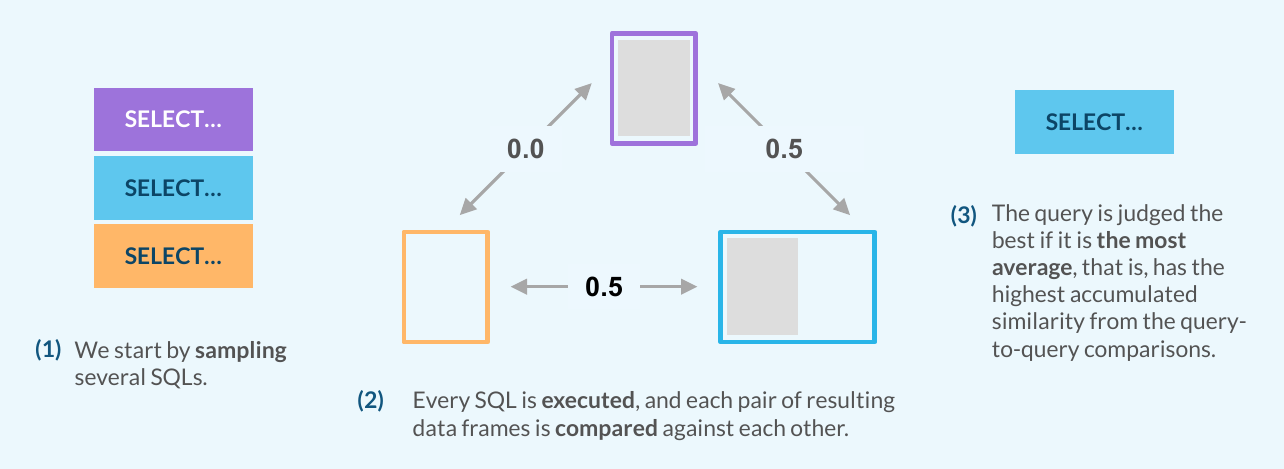}
    \caption{Execution-Guided SQL Generation.}
    \label{fig:hero}
\end{figure*}

Specifically, we propose a novel self-consistency approach tailored to SQL generation, leveraging exact and approximate execution-based similarity metrics to assess semantic equivalence directly from query outputs (Figure~\ref{fig:hero}).
We further frame the problem of self-consistency within the Minimum Bayes Risk (MBR) decoding framework, providing a theoretical justification for our method and extending self-consistency to output spaces defined by execution behavior rather than superficial syntactic forms. 
Finally, we exploit the prefix executability property inherent in specific SQL dialects to incrementally apply execution-based self-consistency during intermediate query generation stages, enabling more robust refinement of complex queries.

These methodological advancements yield substantial empirical improvements. Notably, we demonstrate that by applying execution-based self-consistency, smaller, less expensive models can match the performance of much larger models, highlighting a significant improvement in cost-efficiency.
In particular, the 7B Qwen 2.5 Coder employing our method improves the accuracy by nearly 10\%, reaching the level of O1 despite yielding a 30 times lower inference cost (Figure~\ref{fig:cost}). These findings underscore our method's efficiency, and scalability, positioning it as a strong candidate for real-world SQL generation tasks.

\section{Related Works}

Previous authors addressed SQL generation consistency through heuristics primarily focused on textual and structural similarities among queries \cite{hong2025nextgenerationdatabaseinterfacessurvey,Li_2024,chen2021evaluatinglargelanguagemodels,Li_2024}. Such inherently struggle when queries differ structurally yet produce identical outputs, and thus, are semantically equivalent.

To address this limitation, we propose evaluating equivalence directly at the execution level. Precisely, we execute candidate queries (or approximate their execution) and compare resulting outputs, thus capturing semantic correctness based on actual behavior rather than superficial query structure alone (Figure~\ref{fig:hero}).

Significantly, the proposed method distinguishes itself from recent sequential, multi-step error-correction methods \cite{lee2024mcssqlleveragingmultipleprompts,cen2025sqlfixagentsemanticaccuratetexttosqlparsing,wang2024macsqlmultiagentcollaborativeframework}. These involve iterative refinement, inherently increasing complexity and latency. Conversely, our execution-guided similarity selection enables parallelizable generation processes and relies on efficient comparisons that primarily leverage CPU memory, typically underutilized in standard large language model deployment infrastructures \cite{oliaro2024suffixdecodingmodelfreeapproachspeeding}.

\section{Execution-Guided SQL Generation}

The justification for execution-based similarity stems from the concept of Minimum Bayes Risk decoding \cite{1164173,1164513,kumar-byrne-2004-minimum,eikema-aziz-2020-map}.

\subsection{MBR Decoding and Code Utility}

In MBR decoding, the objective is to find a hypothesis $h^\star$ that minimizes the expected loss (maximizes the expected utility) relative to the set of possible outputs $\mathcal{H}$. If we assume the posterior distribution is derived from the empirical distribution obtained through sampling, it can be described as: \begin{equation*}\begin{split}
    h^\star = \argmax_{h \in \mathcal{H}} \sum_{\hat{h} \in \mathcal{H}} c(\hat{h}) \cdot U(h, \hat{h}) 
\end{split}\end{equation*}
where $c(\hat{h})$ denote the count of a particular hypothesis $\hat{h} \in \mathcal{H}$ in these samples, and $U$ assesses the utility of choosing $h$ when the true hypothesis is $\hat{h}$.

In other words, we select the generation closest on average to all the likely generations, where closeness is measured under some function appropriate for the considered domain.

Concerning majority voting, such utility function is simply: $$U(h, \hat{h}) =
    \begin{cases} 
    1, & \text{if } h = \hat{h}, \\
    0, & \text{otherwise}.
    \end{cases}
$$ and the objective simplifies to: $$h^\star = \argmax_{h \in \mathcal{H}} c(h)$$.

The zero-one utility is justified for short and unambiguous hypotheses, such as numerical answers for mathematical problems faced by LLM, because these are commonly evaluated with exact match accuracy. However, what does it mean for the two distinct SQL statements to be, to some extent, equivalent?

Evaluation practices in the text-to-SQL domain answer this question by resorting to execution-based similarities \cite{zhong2020semanticevaluationtexttosqldistilled}. Thus, the similarity used in self-consistency should be based on the execution equivalence rather than the shallow form of the query or semantic interpretation lacking the information from the execution plan.

\subsection{Execution Similarity}

We follow the abovementioned observation and the notion that comparing two codes' behavior is the natural way to measure their similarity. Such similarity results from yielding the same output under the same circumstances and  naturally captures semantic equivalence, overcoming structural differences in queries \cite{9388651,10.1007/978-3-319-89884-1_11}.

\paragraph{Exact Execution Similarity.} Upon execution, SQL queries we consider return values (cells) structured in horizontal rows and vertical columns identifiable by name.

For the purposes of experiments, we define execution similarity for such \texttt{SELECT} statements based on the resulting tables $A$ and $B$ as a form of recall of cells with respect to the larger column.
Take an example of the following two tables considered column-by-column:

\begin{table}[h]
    \footnotesize
    \centering
    \begin{tabular}{cc}
    \toprule
    X & Y \\
    \midrule
    1 & $\heartsuit$ \\
    2 & $\diamondsuit$ \\
    \bottomrule
    \end{tabular}
    \quad\quad
    \centering
    \begin{tabular}{cc}
    \toprule
    X \\
    \midrule
    1 \\
    2 \\
    3 \\
    \bottomrule
    \end{tabular}
\end{table}

\begin{figure*}[ht]
    \centering
    \includegraphics[width=\linewidth]{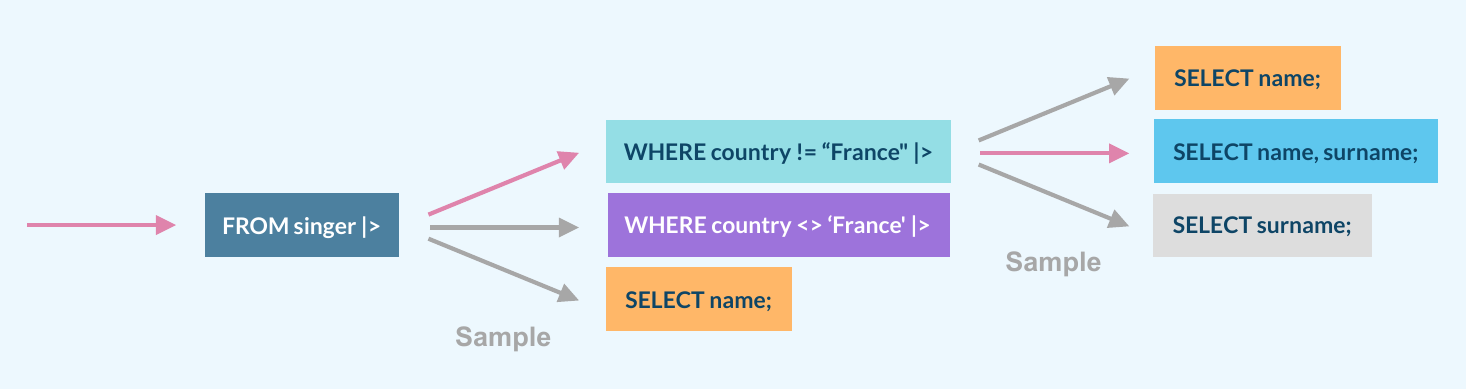}
    \caption{PipeSQL dialect has a property that each query prefix (up to the pipe sequence \texttt{|>}) is also a valid query, making it possible to apply execution-based self-consistency in the middle of the generation process. Instead of sampling $n$ complete SQL sequences, we sample $n$ pipes and stop the generation process. Then, we pick the most consistent pipe and continue the generation sampling $n$ variants of the next pipe.}\label{fig:pipesql}
\end{figure*}

We retrieved two out of three values in column $X$ (1 and 2), and missed two out of two values in column $Y$ ($\heartsuit$, $\diamondsuit$), yielding similarity of $2/5 = 0.4$ (refer to Appendix~\ref{appendix:similarity} for formal definition).

It is one of the many suitable functions that grow monotonically as we approach the desired outcome. Because it relies on the execution result, it fulfills our assumption of measuring behavior. 

\paragraph{Approximate Execution Similarity.} Though in most cases, computing execution similarity is cheap, we consider a variant without the actual execution.
Here, logical execution plans for considered statements are compared, that is, the operations (e.g. table scans and joins) that the database engine would perform to execute the query.

As the execution plan, such as returned by the \texttt{EXPLAIN} query, can be represented as a table, we use the same metric for table-to-table comparison as in the proposed execution similarity. However, this time, it compares the execution plans rather than the actual results.

\subsection{Decoding of Partially-Executable SQL}\label{sec:pipesql}

Parts of standard SQL code, such as subqueries or Common Table Expressions, are executable independently, without the context of the entire query. This property could further improve accuracy without increasing the inference cost, as one could refine the decoding trajectories in the middle of the generation process based on the self-consistency of the considered SQL part.

Due to the property of prefix executability, PipeSQL \cite{pipesql} appears to be even better aligned with this objective. In this dialect, each query prefix (up to the pipe sequence \texttt{|>}) is also a valid query. Consider the example of

\begin{lstlisting}[language=SQL]
    FROM users
    |> WHERE views > 10
    |> ((*\textbf{AGGREGATE}*)) COUNT(id);
\end{lstlisting}

\noindent The first pipe would produce the complete user's table upon execution, the first two pipes---users filtered by the number of views, whereas the complete SQL returns the count of filtered users.

Consequently, one can sample $n$ continuations at each step until the \texttt{|>} sequence and apply execution-based self-consistency on partial completion. Figure~\ref{fig:pipesql} presents an example of such an approach. In the middle of generation, we select a consensus continuation between two \texttt{WHERE} clauses and \texttt{SELECT} based on the similarity matrix:
\begin{figure}[H]
    \centering
    \includegraphics[width=0.35\linewidth]{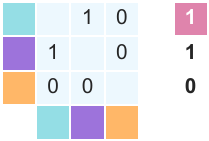}
\end{figure}
There is a tie because \texttt{WHERE} clauses are equivalent, so we pick the first and proceed. In the next step, there are three \texttt{SELECT}s considered:
\begin{figure}[H]
    \centering
    \includegraphics[width=0.35\linewidth]{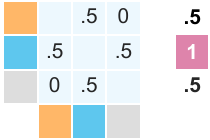}
\end{figure}
The second one is picked---the most average since it produces a union of the remaining ones---and the generation finishes.

Following this idea, we consider text-to-SQL tasks using a partially executable PipeSQL dialect in addition to self-consistency based on complete SQL generation. Such partial-execution-guided decoding can potentially reduce errors early in query generation, improving efficiency and accuracy.

\begin{table*}[ht]
    \renewcommand*{\arraystretch}{1.15}
    \setlength{\tabcolsep}{3.3pt}
    \footnotesize
    \centering
    \begin{tabular}{lc|ccccc|cccccc}
        \toprule
        
         \multirow{2}{*}{\textbf{Model}} & \textbf{Bound} & \multicolumn{3}{c}{\textbf{Baseline Scores}} & \multicolumn{2}{c|}{\textbf{Exec@10}} & \multicolumn{2}{c}{\textbf{Exec@20}} & \multicolumn{2}{c}{\textbf{Exec@30}} \\
         & Pass@10 & Greedy & Maj@10 & Beam@10 & Approx. & Exact & Approx. & Exact & Approx. & Exact \\
         \midrule
         Llama 3.2 3B       & $43.5$ & \cellcolor[HTML]{F5FAFD} {\color{darksnow}18.6} & \cellcolor[HTML]{E5F3FA} {\color{darksnow}20.2} & \cellcolor[HTML]{E7F4FB} {\color{darksnow}20.0} & \cellcolor[HTML]{90CCEC} {\color{darksnow}29.1} & \cellcolor[HTML]{B1DBF2} {\color{darksnow}25.6} & \cellcolor[HTML]{72BEE8} {\color{darksnow}32.2} & \cellcolor[HTML]{75C0E8} {\color{darksnow}31.8} & \cellcolor[HTML]{5BB4E4} {\color{darksnow}34.5} & \cellcolor[HTML]{59B3E4} {\color{darkersnow}34.8} \\
         Qwen 2.5 Coder 3B  & $57.7$ & \cellcolor[HTML]{F5FAFD} {\color{darksnow}30.2} & \cellcolor[HTML]{E1F1FA} {\color{darksnow}32.5} & \cellcolor[HTML]{CBE7F6} {\color{darksnow}34.9} & \cellcolor[HTML]{86C7EB} {\color{darksnow}42.6} & \cellcolor[HTML]{9BD1EE} {\color{darksnow}40.2} & \cellcolor[HTML]{6BBBE6} {\color{darksnow}45.6} & \cellcolor[HTML]{6CBCE7} {\color{darksnow}45.4} & \cellcolor[HTML]{66B9E6} {\color{darksnow}46.1} & \cellcolor[HTML]{59B3E4} {\color{darkersnow}47.6} \\         
         Qwen 2.5 Coder 7B  & $67.9$ & \cellcolor[HTML]{F0F8FC} {\color{darksnow}44.1} & \cellcolor[HTML]{DEF0F9} {\color{darksnow}45.4} & \cellcolor[HTML]{DEF0F9} {\color{darksnow}45.4} & \cellcolor[HTML]{8AC9EC} {\color{darksnow}51.3} & \cellcolor[HTML]{84C7EB} {\color{darksnow}51.7} & \cellcolor[HTML]{78C1E9} {\color{darksnow}52.6} & \cellcolor[HTML]{67B9E6} {\color{darksnow}53.8} & \cellcolor[HTML]{71BEE7} {\color{darksnow}53.1} & \cellcolor[HTML]{59B3E4} {\color{darkersnow}54.8} \\         
         Llama 3.1 8B       & $62.1$ & \cellcolor[HTML]{F5FAFD} {\color{darksnow}32.9} & \cellcolor[HTML]{E7F4FB} {\color{darksnow}34.3} & \cellcolor[HTML]{E1F1FA} {\color{darksnow}34.9} & \cellcolor[HTML]{90CCED} {\color{darksnow}43.1} & \cellcolor[HTML]{8CCAEC} {\color{darksnow}43.6} & \cellcolor[HTML]{82C5EA} {\color{darksnow}44.6} & \cellcolor[HTML]{65B8E6} {\color{darksnow}47.5} & \cellcolor[HTML]{7EC4EA} {\color{darksnow}45.0} & \cellcolor[HTML]{59B3E4} {\color{darkersnow}48.8} \\
         Gemma 3 12B        & $64.9$ & \cellcolor[HTML]{D9EDF8} {\color{darksnow}49.7} & \cellcolor[HTML]{9AD0EE} {\color{darksnow}52.2} & \cellcolor[HTML]{E5F3FA} {\color{darksnow}49.2} & \cellcolor[HTML]{81C5EA} {\color{darksnow}53.2} & \cellcolor[HTML]{86C7EB} {\color{darksnow}53.0} & \cellcolor[HTML]{5BB4E4} {\color{darksnow}54.7} & \cellcolor[HTML]{6FBDE7} {\color{darksnow}53.9} & \cellcolor[HTML]{59B3E4} {\color{darkersnow}54.8} & \cellcolor[HTML]{5EB5E4} {\color{darksnow}54.6} \\
         Qwen 2.5 Coder 14B & $71.5$ & \cellcolor[HTML]{B6DDF3} {\color{darksnow}54.7} & \cellcolor[HTML]{B1DBF2} {\color{darksnow}54.9} & \cellcolor[HTML]{E5F3FA} {\color{darksnow}52.9} & \cellcolor[HTML]{7FC4EA} {\color{darksnow}56.8} & \cellcolor[HTML]{75C0E8} {\color{darksnow}57.2} & \cellcolor[HTML]{70BDE7} {\color{darksnow}57.4} & \cellcolor[HTML]{59B3E4} {\color{darkersnow}58.3} & \cellcolor[HTML]{65B8E6} {\color{darksnow}57.8} & \cellcolor[HTML]{59B3E4} {\color{darkersnow}58.3} \\         
         Codestral 22B {\color{darksnow}\scriptsize v0.1}
                            & $63.5$ & \cellcolor[HTML]{EAF5FB} {\color{darksnow}45.6} & \cellcolor[HTML]{AED9F1} {\color{darksnow}48.6} & \cellcolor[HTML]{DEF0F9} {\color{darksnow}46.2} & \cellcolor[HTML]{85C7EB} {\color{darksnow}50.6} & \cellcolor[HTML]{77C0E8} {\color{darksnow}51.3} & \cellcolor[HTML]{7DC3E9} {\color{darksnow}51.0} & \cellcolor[HTML]{63B7E5} {\color{darksnow}52.3} & \cellcolor[HTML]{71BEE7} {\color{darksnow}51.6} & \cellcolor[HTML]{59B3E4} {\color{darkersnow}52.8} \\
         Gemma 3 27B        & $66.3$ & \cellcolor[HTML]{DAEEF9} {\color{darksnow}53.1} & \cellcolor[HTML]{84C6EB} {\color{darksnow}55.5} & \cellcolor[HTML]{A4D5F0} {\color{darksnow}54.6} & \cellcolor[HTML]{72BEE8} {\color{darksnow}56.0} & \cellcolor[HTML]{80C5EA} {\color{darksnow}55.6} & \cellcolor[HTML]{5CB4E4} {\color{darksnow}56.6} & \cellcolor[HTML]{67B9E6} {\color{darksnow}56.3} & \cellcolor[HTML]{5CB4E4} {\color{darksnow}56.6} & \cellcolor[HTML]{59B3E4} {\color{darkersnow}56.7} \\
         Qwen 2.5 Coder 32B & $71.1$ & \cellcolor[HTML]{D0E9F7} {\color{darksnow}55.0} & \cellcolor[HTML]{C7E5F5} {\color{darksnow}55.2} & \cellcolor[HTML]{D4EBF8} {\color{darksnow}54.9} & \cellcolor[HTML]{98D0EE} {\color{darksnow}56.3} & \cellcolor[HTML]{76C0E8} {\color{darksnow}57.1} & \cellcolor[HTML]{7FC4EA} {\color{darksnow}56.9} & \cellcolor[HTML]{61B6E5} {\color{darksnow}57.6} & \cellcolor[HTML]{59B3E4} {\color{darkersnow}57.8} & \cellcolor[HTML]{61B6E5} {\color{darksnow}57.6} \\
         DeepSeek Coder 33B & $63.4$ & \cellcolor[HTML]{F0F8FC} {\color{darksnow}40.1} & \cellcolor[HTML]{BCE0F4} {\color{darksnow}43.7} & \cellcolor[HTML]{D7EDF8} {\color{darksnow}41.8} & \cellcolor[HTML]{91CDED} {\color{darksnow}46.6} & \cellcolor[HTML]{81C5EA} {\color{darksnow}47.7} & \cellcolor[HTML]{76C0E8} {\color{darksnow}48.5} & \cellcolor[HTML]{64B8E5} {\color{darksnow}49.7} & \cellcolor[HTML]{61B7E5} {\color{darksnow}49.9} & \cellcolor[HTML]{59B3E4} {\color{darkersnow}50.5} \\
         Llama 3.3 70B      & $67.9$ & \cellcolor[HTML]{DBEEF9} {\color{darksnow}53.7} & \cellcolor[HTML]{91CCED} {\color{darksnow}55.8} & \cellcolor[HTML]{B8DEF3} {\color{darksnow}54.7} & \cellcolor[HTML]{86C8EB} {\color{darksnow}56.1} & \cellcolor[HTML]{75BFE8} {\color{darksnow}56.6} & \cellcolor[HTML]{6ABBE6} {\color{darksnow}56.9} & \cellcolor[HTML]{59B3E4} {\color{darkersnow}57.4} & \cellcolor[HTML]{71BEE8} {\color{darksnow}56.7} & \cellcolor[HTML]{60B6E5} {\color{darksnow}57.2} \\
         Mistral Large {\color{darksnow}\scriptsize 2411}
                            & $66.1$ & \cellcolor[HTML]{ACD9F1} {\color{darksnow}53.1} & \cellcolor[HTML]{A6D6F0} {\color{darksnow}53.2} & \cellcolor[HTML]{CBE7F6} {\color{darksnow}52.5} & \cellcolor[HTML]{87C8EB} {\color{darksnow}53.8} & \cellcolor[HTML]{72BEE8} {\color{darksnow}54.2} & \cellcolor[HTML]{8CCAEC} {\color{darksnow}53.7} & \cellcolor[HTML]{5EB5E4} {\color{darksnow}54.6} & \cellcolor[HTML]{7DC3E9} {\color{darksnow}54.0} & \cellcolor[HTML]{59B3E4} {\color{darkersnow}54.7} \\
         Llama 3.1 405 B    & $68.2$ & \cellcolor[HTML]{D6ECF8} {\color{darksnow}54.2} & \cellcolor[HTML]{BEE1F4} {\color{darksnow}54.8} & \na & \cellcolor[HTML]{9DD2EF} {\color{darksnow}55.6} & \cellcolor[HTML]{79C1E9} {\color{darksnow}56.5} & \cellcolor[HTML]{75BFE8} {\color{darksnow}56.6} & \cellcolor[HTML]{59B3E4} {\color{darkersnow}57.3} & \cellcolor[HTML]{71BEE7} {\color{darksnow}56.7} & \cellcolor[HTML]{5DB4E4} {\color{darksnow}57.2} \\
         GPT-4o {\color{darksnow}\scriptsize 2024-11-20}
                            & $62.2$ & \cellcolor[HTML]{B6DDF3} {\color{darksnow}51.6} & \cellcolor[HTML]{B6DDF3} {\color{darksnow}51.6} & \na & \cellcolor[HTML]{B6DDF3} {\color{darksnow}51.6} & \cellcolor[HTML]{7DC3E9} {\color{darksnow}52.4} & \cellcolor[HTML]{7DC3E9} {\color{darksnow}52.4} & \cellcolor[HTML]{6EBCE7} {\color{darksnow}52.6} & \cellcolor[HTML]{6EBCE7} {\color{darksnow}52.6} & \cellcolor[HTML]{59B3E4} {\color{darkersnow}52.9} \\
         GPT-4o mini {\color{darksnow}\scriptsize 2024-11-20}
                            & $62.1$ & \cellcolor[HTML]{E1F1FA} {\color{darksnow}46.9} & \cellcolor[HTML]{9BD1EE} {\color{darksnow}49.3} & \na & \cellcolor[HTML]{79C1E9} {\color{darksnow}50.5} & \cellcolor[HTML]{79C1E9} {\color{darksnow}50.5} & \cellcolor[HTML]{6DBCE7} {\color{darksnow}50.9} & \cellcolor[HTML]{61B7E5} {\color{darksnow}51.3} & \cellcolor[HTML]{64B8E5} {\color{darksnow}51.2} & \cellcolor[HTML]{59B3E4} {\color{darkersnow}51.6} \\
         Gemini 2.0 Flash {\color{darksnow}\scriptsize 001}
                            & $70.9$ & \cellcolor[HTML]{BFE1F4} {\color{darksnow}60.6} & \cellcolor[HTML]{72BEE8} {\color{darksnow}61.8} & \na & \cellcolor[HTML]{78C1E9} {\color{darksnow}61.7} & \cellcolor[HTML]{6CBBE7} {\color{darksnow}61.9} & \cellcolor[HTML]{59B3E4} {\color{darkersnow}62.2} & \cellcolor[HTML]{5FB5E5} {\color{darksnow}62.1} & \cellcolor[HTML]{65B8E6} {\color{darksnow}62.0} & \cellcolor[HTML]{5FB5E5} {\color{darksnow}62.1} \\
         Gemini 2.0 Flash-Lite {\color{darksnow}\scriptsize 02-05}
                            & $69.4$ & \cellcolor[HTML]{D2EAF7} {\color{darksnow}56.5} & \cellcolor[HTML]{93CDED} {\color{darksnow}57.9} & \na & \cellcolor[HTML]{A0D3EF} {\color{darksnow}57.6} & \cellcolor[HTML]{A5D5F0} {\color{darksnow}57.5} & \cellcolor[HTML]{97CFEE} {\color{darksnow}57.8} & \cellcolor[HTML]{6FBDE7} {\color{darksnow}58.7} & \cellcolor[HTML]{93CDED} {\color{darksnow}57.9} & \cellcolor[HTML]{59B3E4} {\color{darkersnow}59.2} \\
         \midrule
         Gemini 2.0 Thinking {\color{darksnow}\scriptsize 01-26} & \na & {\color{darksnow}59.1} & \na & \na & \na & \na & \na & \na & \na & \na \\
         DeepSeek R1 & \na & {\color{darksnow}52.5} & \na & \na & \na & \na & \na & \na & \na & \na \\
         o1 {\color{darksnow}\scriptsize 2024-12-17} & \na & {\color{darksnow}53.9}  & \na & \na & \na & \na & \na & \na & \na & \na \\
         o3-mini {\color{darksnow}\scriptsize 2025-01-31} & \na & {\color{darksnow}52.1}  & \na & \na & \na & \na & \na & \na & \na & \na \\
         \bottomrule
    \end{tabular}
    \caption{BIRD-SQL Accuracy (Text-to-SQLite). The proposed MBR decoding with execution similarity (exec@$n$), compared to baselines: greedy decoding, majority voting with normalization (maj@10), beam search (beam@10), theoretical maximum (pass@10), and heavy reasoning LLMs. Samplings with $\mathrm{temp}=0.7$, validation subset.}    \label{tab:short_results}
\end{table*}

\paragraph{Patience.} In practice, information carried by a single pipe can be too fine-grained for efficient execution-based consistency.
Take an example of join statements that can begin with any of two tables leading to equivalent code when complete SQL sequences are considered:

\begin{lstlisting}[language=SQL]
    FROM emp
    |> JOIN dept ON emp.did = dept.did
    |> SELECT emp.name, dept.name;
\end{lstlisting}

\begin{lstlisting}[language=SQL]
    FROM dept
    |> JOIN emp ON emp.did = dept.did
    |> SELECT emp.name, dept.name;
\end{lstlisting}

The first pipes from each query are dissimilar, even though the sequences of the first and second pairs are not. To accommodate this property, we introduce the patience parameter, which keeps the generation unless it was selected for rejection at least $n$ times in pipe-to-pipe comparisons.

\section{Text-to-SQL Experiments}\label{sec:text2sql}

We perform primary experiments on the BIRD-SQL \cite{li2023llmservedatabaseinterface} dataset. In all cases, the input is the same fixed prompt with the serialized database schema, and the model is requested to generate SQL preceded by a CoT (Appendix~\ref{appendix:prompt}).

A wide range of families and parameter counts is considered, including general-purpose models, such as Llama 3 \cite{grattafiori2024llama3herdmodels}, Gemma 3 \cite{gemma3}, Mistral Large \cite{mistrallarge2}, GPT-4o \cite{openai2024gpt4technicalreport}, and Gemini 2.0 \cite{geminiteam2024geminifamilyhighlycapable}, as well as code-specialized Qwen 2.5 Coder \cite{hui2024qwen25codertechnicalreport}, DeepSeek Coder \cite{guo2024deepseekcoderlargelanguagemodel}, and Codestral \cite{codestral}.

\paragraph{Baselines.} Compute-matched baselines are provided for each model considered in addition to the SQL obtained under greedy decoding. These include the majority vote with SQL normalization\footnote{Using SQLGlot to reformat code: \url{https://github.com/tobymao/sqlglot.git}} and beam search (in fact, beam search with $n$ beams yields even higher cost than sampling $n$ completions, but we assume that they are comparable for simplicity).

\paragraph{Heavy Reasoners.}  Additionally, we provide results of computationally intensive reasoning approaches such as o1 and o3, DeepSeek R1, and Gemini 2.0 Flash Thinking, which require significantly more compute \cite{openai2025competitiveprogramminglargereasoning,deepseekai2025deepseekr1incentivizingreasoningcapability,geminiteam2024geminifamilyhighlycapable}.

\paragraph{Upper bound.} \texttt{Pass@10} score is provided for reference as a theoretical upper bound for all methods based on sampling complete SQL sequences. It is a score with an oracle judge always selecting the best available SQL.

\subsection{Overall Accuracy Improvements}

Results in Table~\ref{tab:short_results} show that execution-guided generation consistently outperforms greedy decoding and compute-matched baselines, holding across a diverse set of language models---from smaller general-purpose LLMs to code-specific and proprietary state-of-the-art systems.

Smaller open-source models (3B--7B parameters) can see substantial boosts, often improving accuracy by 10 points or more with around 30 samples. These improvements highlight that even relatively modest model scales can achieve competitive accuracy when combined with our self-consistency strategy. Larger open-source models, such as Qwen 2.5 Coder 34B, Mistral, or Llama 70B, typically gain 1.5--3.5 points, while huge ones (Llama 405B) still slightly benefit. Closed-source systems also show considerable improvements---GPT-4o mini gains around 5 points, whereas more powerful proprietary models improve by 1--2 points. While improvements for larger models are modest, these results serve primarily as confirmation that applying self-consistency does not negatively impact their already strong performance.

Notably, a 7B Qwen Coder with 30 samples surpasses most heavier reasoning-intensive models, with only Gemini 2.0 Thinking maintaining an edge. Since details on these proprietary systems remain limited, Figure \ref{fig:cost} treats cost as a proxy for inference complexity. It compares Qwen 2.5 Coder 7B to two families of OpenAI models in terms of cost and accuracy, evaluated with and without self-consistency. The results demonstrate that self-consistency significantly increases accuracy at a moderate computational cost, offering an effective balance between quality and resource expenditure.

Although exact variants typically provide superior accuracy, approximate methods remain highly attractive in latency-sensitive or resource-constrained production environments.

Overall, these findings underscore the robust effectiveness of execution-guided generation. Even at more minor model scales, self-consistency offers notable gains in the quality-cost tradeoff, matching or exceeding the performance of more complex proprietary solutions. 

\subsection{Number of Samples and Temperature}

\begin{figure}
    \centering
    \includegraphics[width=0.96\linewidth]{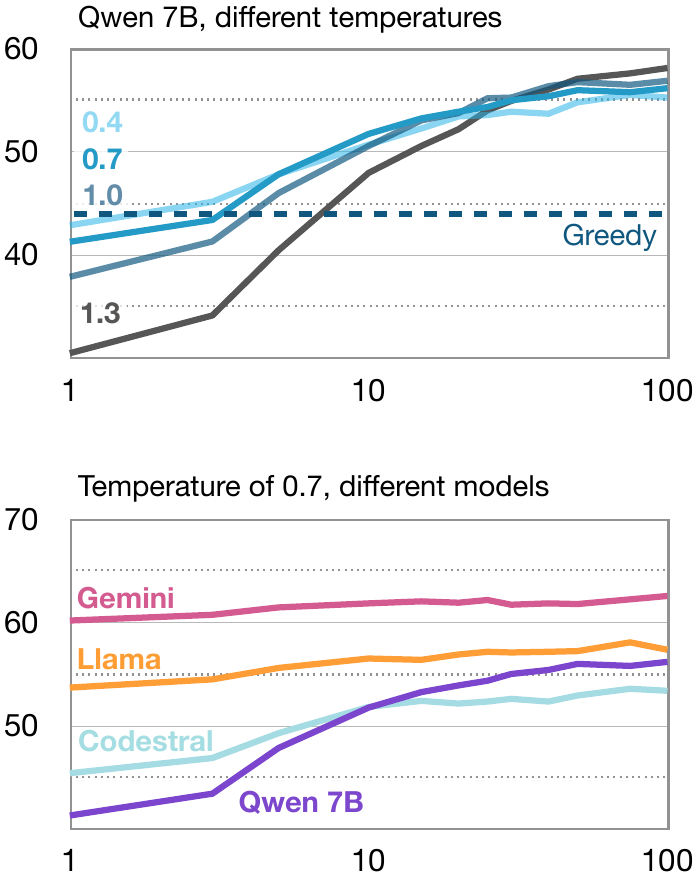}
    \caption{Self-consistency gains for various sample sizes, temperatures, and models (Gemini 2.0 Flash, Llama 3.3 70B, Codestral, Qwen 2.5 Coder 7B).}
    \label{fig:sampling}
\end{figure}

Figure~\ref{fig:sampling} examines how quickly improvements from execution-based self-consistency begin to plateau given different models and temperatures.

Notably, gains become visible with as few as three samples, and using 15 samples proves to be a strong balance between accuracy and computational cost. Although improvements tend to flatten around 50 samples, steady growth continues beyond that point---reflecting how each additional sample helps approximate the distribution of SQL queries generated by the model more accurately.

Sampling temperature is key to balancing immediate accuracy with the potential gains from larger sampling budgets. While increased noise from higher temperatures undermines small-sample performance, it also fosters greater diversity in generated SQLs---ultimately boosting accuracy once the sampling budget is large enough to absorb occasional failures.

In broad terms, weaker models benefit more from the proposed method. 
Interestingly, a high \texttt{pass@1} score does not always translate into superior self-consistency accuracy. E.g., while Qwen Coder 7B outperforms Codestral in a high-sample regime, it underperforms until around 10 samples.

\subsection{Leveraging Partial Executability}\label{sec:partial}

\begin{table}[t]
    \renewcommand*{\arraystretch}{1.15}
    \setlength{\tabcolsep}{6pt}
    \footnotesize
    \centering
    \begin{tabular}{lcccc}
        \toprule
         \textbf{Model} & \textbf{Greedy} & \textbf{Exec} & \textbf{+ Part.} & \textbf{+ Pat.} \\
         \midrule
         Qwen Coder 7B & \cellcolor[HTML]{FAFDFD} {\color{darksnow}27.1} & \cellcolor[HTML]{BCE0E5} {\color{darksnow}41.6} & \cellcolor[HTML]{B7DDE3} {\color{darksnow}42.8} & \cellcolor[HTML]{B1DBE1} {\color{darkersnow}44.3} \\
         Llama 8B & \cellcolor[HTML]{F9FCFC} {\color{darksnow}11.6} & \cellcolor[HTML]{E7F4F6} {\color{darksnow}14.8} & \cellcolor[HTML]{BBDFE5} {\color{darksnow}22.8} & \cellcolor[HTML]{B1DBE1} {\color{darkersnow}24.7} \\
         Gemma 12B & \cellcolor[HTML]{FBFDFD} {\color{darksnow}21.6} & \cellcolor[HTML]{BBDFE5} {\color{darksnow}42.0} & \cellcolor[HTML]{BBDFE5} {\color{darksnow}42.0} & \cellcolor[HTML]{B1DBE1} {\color{darkersnow}45.3} \\
         Qwen Coder 14B & \cellcolor[HTML]{F9FCFD} {\color{darksnow}38.9} & \cellcolor[HTML]{BADFE4} {\color{darksnow}51.2} & \cellcolor[HTML]{C2E3E7} {\color{darksnow}49.6} & \cellcolor[HTML]{B1DBE1} {\color{darkersnow}53.0} \\
         Gemma 27B & \cellcolor[HTML]{FAFDFD} {\color{darksnow}31.2} & \cellcolor[HTML]{B8DEE3} {\color{darksnow}47.3} & \cellcolor[HTML]{BADFE4} {\color{darksnow}46.8} & \cellcolor[HTML]{B1DBE1} {\color{darkersnow}49.1} \\
         Qwen Coder 32B & \cellcolor[HTML]{FAFCFD} {\color{darksnow}40.3} & \cellcolor[HTML]{B7DEE3} {\color{darksnow}53.8} & \cellcolor[HTML]{BADFE4} {\color{darksnow}53.2} & \cellcolor[HTML]{B1DBE1} {\color{darkersnow}55.2} \\
         Codestral & \cellcolor[HTML]{FBFDFD} {\color{darksnow}33.6} & \cellcolor[HTML]{C8E5EA} {\color{darksnow}46.8} & \cellcolor[HTML]{C6E4E9} {\color{darksnow}47.4} & \cellcolor[HTML]{B1DBE1} {\color{darkersnow}53.0} \\
         Llama 70B & \cellcolor[HTML]{FBFDFD} {\color{darksnow}31.3} & \cellcolor[HTML]{B3DCE2} {\color{darksnow}51.2} & \cellcolor[HTML]{BCE0E5} {\color{darksnow}48.7} & \cellcolor[HTML]{B1DBE1} {\color{darkersnow}52.0} \\
         Mistral Large & \cellcolor[HTML]{F6FBFB} {\color{darksnow}44.3} & \cellcolor[HTML]{C5E4E9} {\color{darksnow}50.4} & \cellcolor[HTML]{C2E3E7} {\color{darksnow}50.8} & \cellcolor[HTML]{B1DBE1} {\color{darkersnow}53.0} \\ %
         LLama 405B & \cellcolor[HTML]{FBFDFD} {\color{darksnow}37.4} & \cellcolor[HTML]{BBDFE4} {\color{darksnow}54.0} & \cellcolor[HTML]{BDE0E5} {\color{darksnow}53.4} & \cellcolor[HTML]{B1DBE1} {\color{darkersnow}56.7} \\
         \bottomrule
    \end{tabular}
    \caption{BIRD-SQL Accuracy (Text-to-PipeSQL). Greedy decoding results compared to ten samples budged with standard execution-based self-consistency, partial self-consistency, or its variant with patience.}
    \label{tab:pipesql}
\end{table}


Since we rely on the PipeSQL dialect for partial executability (see~Section\ref{sec:pipesql}), these experiments required transpiling ground-truth BIRD-SQL queries, converting the underlying databases, and establishing separate baseline scores (Appendix~\ref{appendix:pipebird}).

Moreover, unlike the widely used SQLite dialect in previous experiments, PipeSQL is relatively novel and not recognized by default, necessitating thorough in-prompt documentation with examples. Consequently, our inputs in this setup reached approximately 15k tokens, so certain open-source models with limited context windows could not be evaluated. Commercial API-based models were similarly excluded, as their interfaces typically do not allow mid-generation refinement.

Despite few-shot prompting and including detailed documentation on the new dialect, models typically perform substantially worse with PipeSQL compared to standard SQL. Switching from SQLite yields accuracy drops of 5--20 points depending on model capabilities, underscoring limited generalizability to novel query dialects and problems with instruction-following.

Table~\ref{tab:pipesql} summarizes the results in multiple settings: (1) Greedy decoding, (2) Standard self-consistency without leveraging partial executability, (3) Self-consistency enhanced by partial, pipe-by-pipe executability, and (4) Partial executability with a patience parameter ($n=3$), accommodating temporary divergences in intermediate SQL steps.

\begin{figure*}[t]
    \begin{minipage}{.475\textwidth} 
    \centering
    \includegraphics[width=\linewidth]{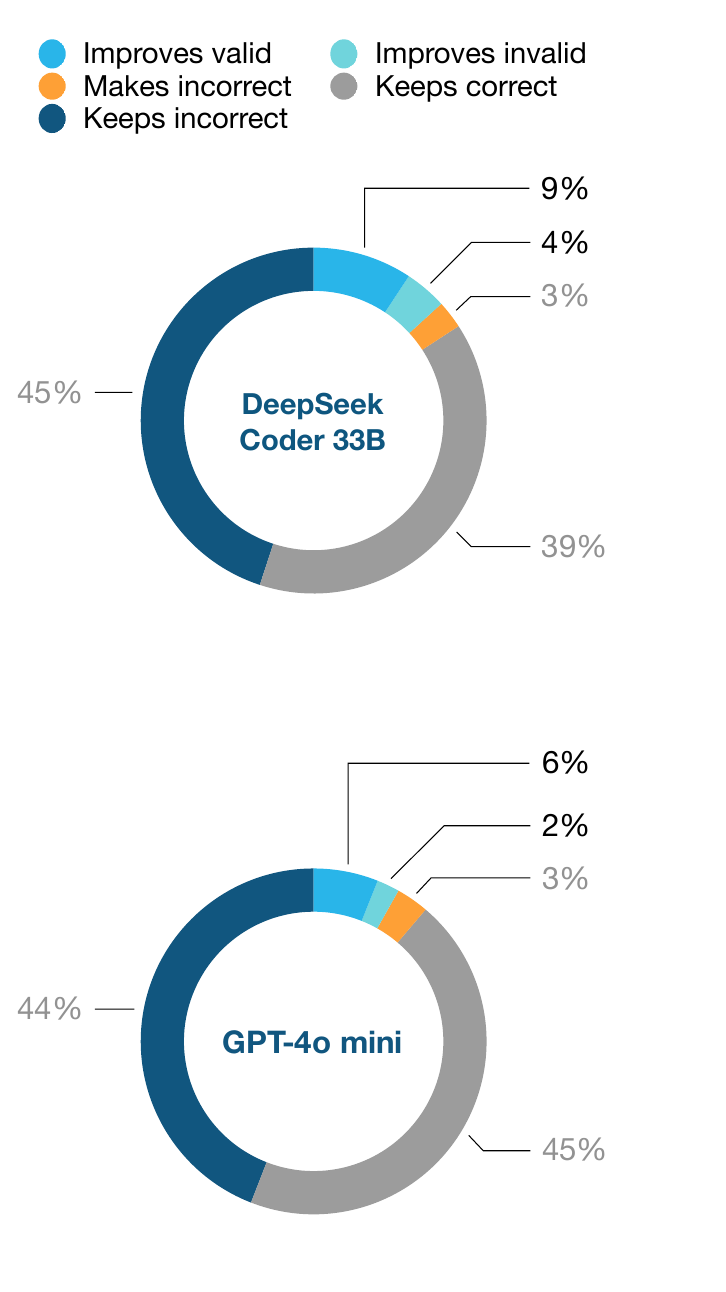}
    \caption{Effect of replacing outputs produced under greedy decoding by self-consistency outputs. \textit{Valid} and \textit{invalid} refer to executability, whereas \textit{correct} and \textit{incorrect}---conforming to the gold standard.}
    \label{fig:types}
    \end{minipage}\hspace{.05\textwidth}
    \begin{minipage}{.475\textwidth}
    \centering
    \includegraphics[width=0.995\linewidth]{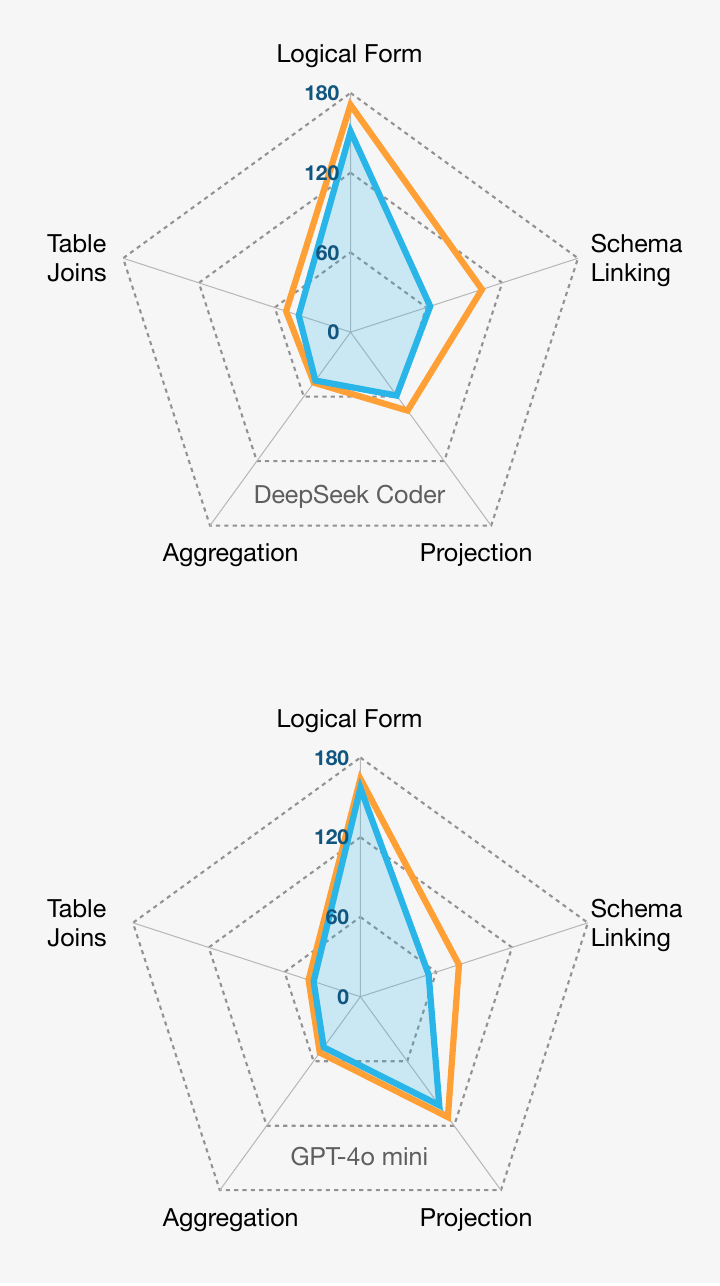}
    \caption{Top problems explaining why BIRD-SQL generations of DeepSeek Coder and GPT-4o mini were incorrect. Greedy decoding compared to self-consistency outputs.}
    \label{fig:taxonomy}
    \end{minipage}
\end{figure*}

Standard self-consistency notably improves accuracy over greedy decoding (typically by around 15 points), because non-executable or failing SQL generations, more common with unfamiliar dialects, naturally receive low similarity scores and are thus filtered out.

Partial pipe-by-pipe executability alone provides mixed results. Depending on the model, accuracy may slightly increase or decrease (1--2 points) compared to standard self-consistency. Finally, introducing the patience parameter generally yields further accuracy gains (1--10 points over standard self-consistency), demonstrating the benefit of tolerating intermediate divergences.

We hypothesize considerable potential in the partial executability approach but fully realizing it requires models capable of generating high-quality PipeSQL. Currently, limitations include insufficient training data on such dialects and the lack of robust transpilers for existing SQL datasets.

\section{Qualitative Analysis}\label{sec:analysis}

To gain deeper insights into how execution-based self-consistency enhances text-to-SQL generation, we analyze the specific error types it effectively addresses and present key statistical comparisons between greedy-decoded outputs and those refined through self-consistency.

The starting point for this part is the previously established taxonomy of SQL generation errors \cite{shen2025studyincontextlearningbasedtexttosqlerrors} and linguistic ambiguities in text-to-SQL tasks \cite{huang2023dataambiguitystrikesback}. Specifically, we consider common categories of mistakes, including
dialect mismatches, schema linking failures, data type mismatches, incorrect aggregation, logical form inaccuracies, improper table joins, and projection errors (see Appendix~\ref{appendix:taxonomy} for details).

Figure~\ref{fig:types} examines the per-example impact of adopting self-consistency compared to greedy decoding. While self-consistency occasionally yields inferior SQL queries, its overall effect is beneficial. Most improvements occur because the method successfully replaces executable but incorrect queries with queries that correctly address the user’s intent.

A closer inspection of the error categories reveals that most text-to-SQL failures stem from flawed logical forms, schema linking mistakes, and incorrect projections. These arise when queries are not logically coherent when the model confuses or hallucinates database elements (e.g., referencing nonexistent columns), and when it selects columns that do not match the intent of the user query.

Execution-based self-consistency mitigates these issues by filtering out candidates that fail under execution or yield outlier data frames, thus promoting solutions that produce similar outputs across multiple independent samples. For example, applying self-consistency in DeepSeek Coder reduces schema linking errors by 40\%, projection, and table join mistakes by 20\%, and logical form errors by 11\% (see Figure~\ref{fig:taxonomy}). Models with fewer initial errors tend to benefit less; for instance, GPT-4o mini still sees a notable 30\% drop in schema linking mistakes. However, the low incidence of such errors in its greedy baseline limits the overall improvement.

Hence, this error-reduction pattern also aligns with the findings described in the earlier sections, where models with lower baseline performance benefited disproportionately more from execution-based self-consistency.

\section{Limitations}

While our self-consistency provides substantial accuracy gains, it requires generating multiple candidate queries and assessing their behavior. In some cases, actual query execution can be expensive or time-consuming. Still, the proposed approximate approach significantly reduces this concern since \texttt{EXPLAIN} queries typically incur negligible costs.

Moreover, generating multiple solutions introduces additional compute time compared to a single pass, yet the approach remains easily parallelizable. By contrast, step-by-step agentic solutions can inflate latency significantly because their sequential reasoning is far more challenging to distribute across multiple processes.

Next, even though our preliminary evaluation with the partially executable SQL dialect suggests promising results, a direct comparison to standard SQLite tasks remains an area for further studies.

Finally, models with fewer initial errors or a high baseline accuracy may exhibit diminishing returns, limiting the offered improvement.

\section{Summary}
A family of self-consistency techniques relying on sampling multiple SQL queries and comparing their execution results has been introduced. 
Additionally, a partial-execution variant allowing step-by-step refinement of intermediate query fragments was proposed, enabling further precision improvements when partial queries are reliably generated.

The proposed methods robustly identify semantically equivalent queries even when there are structural variations, allowing smaller and inexpensive models to achieve accuracy typically reserved for larger and costlier LLMs
---all while avoiding substantial processing time overhead typically associated with iterative refinement strategies. 

The presented analysis reveals that offered improvements can be attributed to effectively addressing common SQL generation errors, yielding 20--40\% reductions in schema linking errors, projection and table join mistakes, and logical form errors. 

While the presented experiments focused on SQL generation, the underlying principle of leveraging execution-based similarity naturally extends to other programming languages and code-generation tasks. Supplementary results on text-to-code benchmarks provided in Appendix~\ref{appendix:python} suggest the broad applicability of execution-guided self-consistency and highlight its potential in program synthesis across diverse domains.

We believe that further exploration of execution-guided generation will open promising avenues toward robust, efficient, and universally applicable code-generation models.

\section*{Acknowledgments}
We sincerely thank the colleagues whose insightful discussions and feedback shaped this paper.
Special thanks to Anupam Datta, Michał Zając, Filip Graliński, Zhewei Yao, Andrzej Szwabe, Michał Pietruszka, Jakub Świątkowski, Wojciech Jaśkowski, and Tomasz Dwojak for generously sharing their perspectives, inspiring research questions, and impacting the evolution of our ideas.
\clearpage
\bibliography{custom}

\appendix

\section{Supplementary Experiments}
\subsection{Beyond SQL Generation}\label{appendix:python}

The execution-based self-consistency we proposed has applications beyond SQL generation, e.g., for ordinary Python code.

We consider two functions similar if they produce the same outputs for the same inputs. The more inputs we test and outputs we compare, the more we know about their similarity. For example, we would expect a similarity of $0.5$ with Python functions:

\begin{lstlisting}[language=Python]
    def A(x: int) -> int:
        return x
    
    def B(x: int) -> int:
        return x if x % 2 == 0 else 0
\end{lstlisting}

\noindent as they yield the same output for half of the inputs. Specifically, given functions $A, B: X \mapsto Y$ we sample arguments $\{x_1, \dots, x_n\}$ and measure the expected fraction of agreement between $A(x)$ and $B(x)$ over the domain $X$. Similarity is defined as:
$$S(A, B) = \frac{1}{n} \sum_{i=1}^n \mathbf{1}[A(x_i) = B(x_i)],$$ where $\mathbf{1}[A(x_i) = B(x_i)]$ is the indicator function that checks if $A(x_i)$ equals $B(x_i)$.

\paragraph{Experiments.}
In practice, using random input values is sample-inefficient, and having a source of somewhat reasonable inputs concerning the intended behavior is better.

For simplicity, in this section, we assume such a source is given and rely on real test cases from HumanEval+ and MBPP+ benchmarks \cite{evalplus}. Note that we do not leverage function definitions or expected outputs but merely the tested function arguments.

\begin{table}[t]
    \renewcommand*{\arraystretch}{1.13}
    \setlength{\tabcolsep}{3.3pt}
    \footnotesize
    \centering
    \subfloat[HumanEval+]{%
        \begin{tabular}{lcc}
        \toprule
        \textbf{Model} & \textbf{Greedy} & \textbf{Exec@10} \\
        \midrule
        Mistral Large {\color{darksnow}\scriptsize 2407} & \cellcolor[HTML]{B2DBE1} {\color{darksnow}84.8} & \cellcolor[HTML]{B1DBE1} {\color{darkersnow}85.2} \\
        Codestral 22B & \cellcolor[HTML]{D5EBEE} {\color{darksnow}73.8} & \cellcolor[HTML]{C5E4E9} {\color{darksnow}78.6} \\
        Llama 3.1 8B & \cellcolor[HTML]{FBFDFD} {\color{darksnow}61.6} & \cellcolor[HTML]{EDF6F8} {\color{darksnow}66.1} \\
        Llama 3.1 405B & \cellcolor[HTML]{C1E2E7} {\color{darksnow}79.9} & \cellcolor[HTML]{B5DDE2} {\color{darksnow}83.8} \\
        \bottomrule
        \end{tabular}
    }
    \vspace{1em}
    \subfloat[MBPP+]{%
        \begin{tabular}{lcc}
        \toprule
        \textbf{Model} & \textbf{Greedy} & \textbf{Exec@10} \\
        \midrule
        Mistral Large {\color{darksnow}\scriptsize 2407} & \cellcolor[HTML]{D7EDF0} {\color{darksnow}65.9} & \cellcolor[HTML]{B8DEE3} {\color{darksnow}75.3} \\
        Codestral 22B & \cellcolor[HTML]{E3F2F4} {\color{darksnow}62.4} & \cellcolor[HTML]{C2E2E7} {\color{darksnow}72.4} \\
        Llama 3.1 8B & \cellcolor[HTML]{FBFDFD} {\color{darksnow}55.3} & \cellcolor[HTML]{D0E9ED} {\color{darksnow}68.0} \\
        Llama 3.1 405B & \cellcolor[HTML]{C8E6EA} {\color{darksnow}70.4} & \cellcolor[HTML]{B1DBE1} {\color{darkersnow}77.5} \\
        \bottomrule
        \end{tabular}
    }
    \caption{Python generation accuracy. Greedy decoding compared to execution-based self-consistency.}
    \label{tab:python}
\end{table}

We observe consistent accuracy gains under execution-based self-consistency in both HumanEval+ and MBPP+ (Table~\ref{tab:python}). On HumanEval+, the improvements tend to be modest---e.g., Mistral Large rises from 84.8\% to 85.2\%, while Llama 3.1~8B increases by nearly five points (61.6\% to 66.1\%). In contrast, MBPP+ shows more pronounced gains: Mistral Large jumps by almost ten points (65.9\% to 75.3\%), and Llama 3.1~8B improves by over twelve (55.3\% to 68.0\%).

These results indicate that execution-based comparisons become increasingly valuable for more challenging code generation benchmarks, where model outputs show more significant variability, and there is more room for error correction by filtering out inconsistent or failing candidates.

\subsection{Cross-Model SQL Consistency}

Another way to broaden the range of generated solutions is to sample from multiple LLMs rather than relying on a single one.

Results presented in Table~\ref{tab:cross} indicate that combining samples from multiple LLMs, even without weighting them by expected accuracy, yields modest but consistent accuracy improvements---typically on the order of 0.5 to 1 point. In the larger model group, for instance, pairing a stronger model (Qwen Coder 32B) with a weaker one (Llama 70B) provides a slight boost over using only the stronger model for the same total number of samples. Adding the more capable Gemini model to this mix further improves results.

Similar trends emerge among smaller and mid-sized models (e.g. Qwen 2.5 Coder 7B, Deepseek Coder, and Codestral), where combining their outputs slightly enhances performance without increasing the total sample count in self-consistency.

Despite these incremental gains, the approach is appealing for its simplicity and low overhead, underscoring the potential of straightforward cross-model ensembling.

\begin{table}[t]
    \renewcommand*{\arraystretch}{1.3}
    \setlength{\tabcolsep}{4pt}
    \footnotesize
    \centering
    \begin{tabular}{lc|ccc}
        \toprule
         \multirow{2}{*}{\textbf{Model}} & \textbf{Bound} & \multicolumn{3}{c}{\textbf{Sample Budget}} \\
         & Pass@10 & @10 & @20 & @30 \\
         \midrule
         DeepSeek Coder 33B & $63.4$ & \cellcolor[HTML]{F6FBFB} {\color{darksnow}47.7} & \cellcolor[HTML]{E5F3F5} {\color{darksnow}49.7} & \cellcolor[HTML]{DEF0F2} {\color{darksnow}50.5} \\
         Qwen Coder 7B & $67.9$ & \cellcolor[HTML]{D4EBEE} {\color{darksnow}51.7} & \cellcolor[HTML]{C2E3E7} {\color{darksnow}53.8} & \cellcolor[HTML]{BADFE4} {\color{darksnow}54.8} \\ 
         Codestral 22B {\color{darksnow}\scriptsize v0.1} & $63.5$ & \cellcolor[HTML]{D8EDF0} {\color{darksnow}51.3} & \cellcolor[HTML]{CFE9EC} {\color{darksnow}52.3} & \cellcolor[HTML]{CBE7EB} {\color{darksnow}52.8} \\
         \quad + DeepSeek Coder 33B & $66.5$ & \cellcolor[HTML]{D7ECEF} {\color{darksnow}51.4} & \cellcolor[HTML]{C7E5E9} {\color{darksnow}53.3} & \cellcolor[HTML]{BFE1E6} {\color{darksnow}54.2} \\
         \quad \quad + Qwen Coder 7B & $70.1$ & \cellcolor[HTML]{CCE7EB} {\color{darksnow}52.7} & \cellcolor[HTML]{B9DEE4} {\color{darksnow}54.9} & \cellcolor[HTML]{B1DBE1} {\color{darkersnow}55.9} \\
         \midrule
 Llama 70B & $67.9$ & \cellcolor[HTML]{F3F9FA} {\color{darksnow}56.6} & \cellcolor[HTML]{EAF5F7} {\color{darksnow}57.4} & \cellcolor[HTML]{EDF6F8} {\color{darksnow}57.2} \\
         Gemini Flash {\color{darksnow}\scriptsize 001} & $70.9$ & \cellcolor[HTML]{B8DEE4} {\color{darksnow}61.9} & \cellcolor[HTML]{B6DDE3} {\color{darksnow}62.1} & \cellcolor[HTML]{B6DDE3} {\color{darksnow}62.1} \\
         Qwen Coder 32B & $71.1$ & \cellcolor[HTML]{EEF7F8} {\color{darksnow}57.1} & \cellcolor[HTML]{E8F4F6} {\color{darksnow}57.6} & \cellcolor[HTML]{E8F4F6} {\color{darksnow}57.6} \\
         \quad + Llama 70B & $72.2$ & \cellcolor[HTML]{EAF5F7} {\color{darksnow}57.4} & \cellcolor[HTML]{E3F2F4} {\color{darksnow}58.1} & \cellcolor[HTML]{DFF0F3} {\color{darksnow}58.4} \\
         \quad\quad + Gemini Flash & $74.6$ & \cellcolor[HTML]{BFE1E6} {\color{darksnow}61.3} & \cellcolor[HTML]{B6DDE3} {\color{darksnow}62.1} & \cellcolor[HTML]{B1DBE1} {\color{darkersnow}62.6} \\
         \bottomrule
    \end{tabular}
    \caption{Impact of cross-model consistency on BIRD-SQL Accuracy (Text-to-SQLite).}
    \label{tab:cross}
\end{table}




\section{Details of SQL Experiments}

All open-source language models were evaluated using \texttt{vLLM} \cite{kwon2023efficient}, except for the beam-search baselines, which relied on the Transformers library \cite{wolf-etal-2020-transformers}.
In both setups, inference runs were performed on Nvidia DGX nodes with 8$\times$H100 GPUs. Nearly all inferences relied on \texttt{bf16} precision, except Llama 405B, which was evaluated in \texttt{fp8} due to hardware constraints.

OpenAI models were accessed through their public API, Gemini models through the proprietary Google interface, and DeepSeek R1 via Snowflake Cortex. When reasoning-intensive models were examined, up to 32k tokens were allowed for their chain-of-thought processes.

\subsection{Data Frame Similarity}\label{appendix:similarity}

Specifically, we define similarity as: $$S(A, B) = \frac{R}{\max\left(|A|, |B|\right)}$$
$$R = \sum_{c \in C} \sum_{v \in V_c} \min\big(f_{A_c}(v), f_{B_c}(v)\big),$$ where $|A|$ and $|B|$ denote the total number of cells in tables, $C$ is the set of all column names, $V_c$  is the set of unique values in column $c$, whereas $f_{A_c}(v)$ and $f_{B_c}(v)$ denote the frequency of a value $v$ in column $c$ of $A$ and $B$, respectively.

\subsection{Inference Cost}

All cost estimates reflect actual billing under non-batch inference. Specifically, OpenAI’s usage-based pricing was applied to the OpenAI models, while Together.ai’s pricing was used to obtain Qwen 2.5 Coder costs.
Because the Qwen 2.5 Coder 7B was unavailable on Together.ai, its cost was approximated based on the non-coder Qwen 2.5 7B Instruct Turbo pricing tier.

\subsection{BIRD for PipeSQL}\label{appendix:pipebird}

For simplicity, gold standard queries were converted to the regular BigQuery format rather than an explicit pipe-based syntax. Since the same execution engine powers both dialects and produces identical results for semantically equivalent queries, the choice does not affect correctness.

Each original query was first transpiled from SQLite to BigQuery using the SQLGlot library. Both forms were executed---one in SQLite, the other in BigQuery---and their resulting data frames were compared for consistency. If the SQLGlot-transpiled query failed or did not match the original query’s result, the query was processed by an LLM (o3-mini) for manual repair. The corrected query was again tested for matching data frames.

For fewer than 100 instances, neither automatic nor LLM-based transpilation yielded matching results. These queries were removed from the final PipeSQL dataset. 
All other queries were successfully converted, ensuring that the PipeSQL variant of BIRD-SQL closely mirrors the original in both content and accuracy.

\subsection{SQL Error Taxonomy}\label{appendix:taxonomy}

For the purposes of qualitative analysis, we simplify the previously established taxonomy of SQL generation errors \cite{shen2025studyincontextlearningbasedtexttosqlerrors} and linguistic ambiguities in text-to-SQL tasks \cite{huang2023dataambiguitystrikesback} distinguishing the following categories.

\paragraph{Dialect.} Errors arising from dialect differences between SQL variants. A generated query might be semantically and syntactically correct in general but fail when executed due to differences specific to the target database.

\paragraph{Schema Linking.} Errors due to incorrect or failed mappings between natural language references and corresponding database schema elements. This includes hallucinating non-existent tables or columns that appear relevant.

\paragraph{Data Type.} Errors occurring when queries fail or produce incorrect results due to mismatches or unexpected data types within the database.

\paragraph{Aggregation.} Errors involving incorrect use or omission of aggregation functions (e.g. \texttt{COUNT}, \texttt{SUM}), leading to inaccurate summary calculations or improperly grouped results.

\paragraph{Logical Form and Condition.} Errors resulting from incorrect logical structures or incomplete query specifications. This includes missing or incorrect conditions, inappropriate filters, erroneous ordering logic, or improperly scoped constraints.

\paragraph{Table Joins.} Errors involving incorrect table relationships, such as joining irrelevant tables, omitting necessary joins, or misidentifying join conditions.

\paragraph{Projection.} The model may choose the wrong columns or expressions to return. Sometimes, it might select a computed value when the question asks for an entity or vice versa.

\lstset{
  basicstyle=\ttfamily\small,
  columns=fullflexible,
  breaklines=true,
  postbreak=\mbox{\textcolor{dark}{$\hookrightarrow$}\space},
  backgroundcolor=\color{backcolour2},   
  framexleftmargin=6pt,
  framexrightmargin=6pt,
  framextopmargin=6pt,
  framexbottommargin=6pt,
  frame=tb, framerule=0.7pt,
  xleftmargin=0.3cm,
}

\section{Prompts}
\subsection{SQLite Generation}\label{appendix:prompt}
For BIRD-SQL, we provide a straightforward instruction, serialised database, and questions concatenated with evidence.

\begin{lstlisting}
You are an AI assistant helping a data analyst write SQL queries to answer questions. Below I will provide a DB schema with example values and a question that can be answered by querying the provided DB. You will then write out your thought process in detail followed by a single SQL query enclosed in ```sql ...``` that answers the question.

SQLite database schema:
Table: alignment :
id : integer, primary key, example values: ( 1 , 2 , 3 )
alignment : text, example values: ( 'Good' , 'Bad' , 'Neutral' )

Table: attribute :
id : integer, primary key, example values: ( 1 , 2 , 3 )
attribute_name : text, example values: ( 'Intelligence' , 'Strength' , 'Speed' )

Table: colour :
id : integer, primary key, example values: ( 1 , 2 , 3 )
colour : text, example values: ( 'No Colour' , 'Amber' , 'Auburn' )

Table: gender :
id : integer, primary key, example values: ( 1 , 2 , 3 )
gender : text, example values: ( 'Male' , 'Female' , 'N/A' )

Table: publisher :
id : integer, primary key, example values: ( 1 , 2 , 3 )
publisher_name : text, example values: ( '' , 'ABC Studios' , 'Dark Horse Comics' )

Table: race :
id : integer, primary key, example values: ( 1 , 2 , 3 )
race : text, example values: ( '-' , 'Alien' , 'Alpha' )

Table: superhero :
id : integer, primary key, example values: ( 1 , 2 , 3 )
superhero_name : text, example values: ( '3-D Man' , 'A-Bomb' , 'Abe Sapien' )
full_name : text, example values: ( 'Charles Chandler' , 'Richard Milhouse Jones' , 'Abraham Sapien' )
gender_id : integer, foreign key, references gender, example values: ( 1 , 2 , 3 )
eye_colour_id : integer, foreign key, references colour, example values: ( 9 , 33 , 7 )
hair_colour_id : integer, foreign key, references colour, example values: ( 13 , 1 , 4 )
skin_colour_id : integer, foreign key, references colour, example values: ( 1 , 7 , 23 )
race_id : integer, foreign key, references race, example values: ( 1 , 24 , 33 )
publisher_id : integer, foreign key, references publisher, example values: ( 13 , 3 , 4 )
alignment_id : integer, foreign key, references alignment, example values: ( 1 , 2 )
height_cm : integer, example values: ( 188 , 203 , 191 )
weight_kg : integer, example values: ( 90 , 441 , 65 )

Table: hero_attribute :
hero_id : integer, foreign key, references superhero, example values: ( 1 , 2 , 3 )
attribute_id : integer, foreign key, references attribute, example values: ( 1 , 2 , 3 )
attribute_value : integer, example values: ( 80 , 75 , 95 )

Table: superpower :
id : integer, primary key, example values: ( 1 , 2 , 3 )
power_name : text, example values: ( 'Agility' , 'Accelerated Healing' , 'Lantern Power Ring' )

Table: hero_power :
hero_id : integer, foreign key, references superhero, example values: ( 1 , 2 , 3 )
power_id : integer, foreign key, references superpower, example values: ( 1 , 18 , 26 )

The question: Who is the publisher of Sauron? (the publisher refers to publisher_name; Sauron refers to superhero_name = 'Sauron')"
\end{lstlisting}

\lstset{
  backgroundcolor=\color{backcolour},
}

\subsection{PipeSQL Generation}\label{appendix:pipe_prompt}
For PipeSQL generation, we provide quite elaborate prompt. It consists of simplified dialect documentation\footnote{\url{https://github.com/google/zetasql/blob/master/docs/pipe-syntax.md}} with complex examples of transpiled TPC-H queries\footnote{\url{https://github.com/google/zetasql/tree/master/zetasql/examples/tpch/pipe_queries}} \cite{tpc-h-2.17.1}.
\begin{lstlisting}
You will be asked to generate SQL using BigQuery's pipe query syntax.

# Pipe Query Syntax
Pipe syntax has the following key characteristics
- Each pipe operator in pipe syntax consists of the pipe symbol, |>, an operator name, and any arguments.
- Pipe syntax works anywhere standard syntax is supported: in queries, views, table-valued functions (TVFs), and other contexts.
- Pipe syntax can be mixed with standard syntax in the same query. E.g., subqueries can use different syntax from the parent query.
- A query can start with a FROM clause, and pipe operators can optionally be added after the FROM clause.

Pipe operators have the following semantic behavior
- Each pipe operator performs a self-contained operation.
- A pipe operator consumes the input table passed to it through the pipe symbol, |>, and produces a new table as output.
- A pipe operator can reference only columns from its immediate input table. Columns from earlier in the same query aren't visible. Inside subqueries, correlated references to outer columns are still allowed.

Operator list
- SELECT: Produces a new table with the listed columns.
- EXTEND: Propagates the existing table and adds computed columns.
- SET:	Replaces the values of columns in the current table.
- DROP: Removes listed columns from the current table.
- RENAME: Renames specified columns.
- AS: Introduces a table alias for the input table.
- WHERE: Filters the results of the input table.
- LIMIT: Limits the number of rows to return in a query, with an optional  OFFSET clause to skip over rows.
- AGGREGATE: Performs aggregation on data across groups of rows or the full input table.
- DISTINCT: Returns distinct rows from the input table, while preserving table aliases.
- ORDER BY: Sorts results by a list of expressions.
- UNION: Combines the results of the input queries to the left and right of the pipe operator by pairing columns from the results of each query and vertically concatenating them.
- INTERSECT: Returns rows that are found in the results of both the input query to the left of the pipe operator and all input queries to the right of the pipe operator.
- EXCEPT: Returns rows from the input query to the left of the pipe operator that aren't present in any input queries to the right of the pipe operator.
- JOIN: Joins rows from the input table with rows from a second table provided as an argument.
- CALL: Calls a table-valued function (TVF), passing the pipe input table as a table argument.
- WINDOW: Adds columns with the result of computing the function over some window of existing rows
- TABLESAMPLE: Selects a random sample of rows from the input table.
- PIVOT: Rotates rows into columns.
- UNPIVOT: Rotates columns into rows.
- ASSERT: Evaluates that an expression is true for all input rows, raising an error if not.

## Examples
### 1. Pricing Summary Report Query
How can I generate a pricing summary report that shows, for each combination of return flag and line status, the total quantity shipped, total base price, total discounted price, total charge (discounted price plus tax), average quantity, average extended price, average discount, and the count of orders? The report should only include line items shipped on or before the date obtained by subtracting 74 days from December 1, 1998, and it should be ordered by return flag and line status in ascending order.

```sql
SELECT
  l_returnflag,
  l_linestatus,
  sum(l_quantity) AS sum_qty,
  sum(l_extendedprice) AS sum_base_price,
  sum(l_extendedprice * (1 - l_discount)) AS sum_disc_price,
  sum(l_extendedprice * (1 - l_discount) * (1 + l_tax)) AS sum_charge,
  avg(l_quantity) AS avg_qty,
  avg(l_extendedprice) AS avg_price,
  avg(l_discount) AS avg_disc,
  COUNT(*) AS count_order
FROM
  lineitem
WHERE
  l_shipdate <= date_sub(date '1998-12-01', INTERVAL 74 day)
GROUP BY
  l_returnflag,
  l_linestatus
ORDER BY
  l_returnflag,
  l_linestatus;
```

### 2. Minimum Cost Supplier Query
Which supplier in the Middle East should I select to order parts of size 19 that are of a type containing 'COPPER', based on the lowest available supply cost? If multiple suppliers offer the part at the same minimum cost, I want to consider only the top 100 suppliers with the highest account balances. For each supplier, please provide their account balance, name, nation, the part number, manufacturer, address, phone number, and any additional comments, and sort the results by account balance (highest first), then by nation, supplier name, and part number.

```sql
FROM
  part,
  supplier,
  partsupp,
  nation,
  region
|> WHERE
     p_partkey = ps_partkey
     AND s_suppkey = ps_suppkey
     AND p_size = 19
     AND p_type LIKE '%COPPER'
     AND s_nationkey = n_nationkey
     AND n_regionkey = r_regionkey
     AND r_name = 'MIDDLE EAST'
     AND ps_supplycost = (
       FROM
         partsupp,
         supplier,
         nation,
         region
       |> WHERE
            p_partkey = ps_partkey
            AND s_suppkey = ps_suppkey
            AND s_nationkey = n_nationkey
            AND n_regionkey = r_regionkey
            AND r_name = 'MIDDLE EAST'
       |> AGGREGATE
            min(ps_supplycost))
|> SELECT
     s_acctbal,
     s_name,
     n_name,
     p_partkey,
     p_mfgr,
     s_address,
     s_phone,
     s_comment
|> ORDER BY
     s_acctbal DESC,
     n_name,
     s_name,
     p_partkey
|> LIMIT 100;
```

### 3. Order Priority Checking Query
Can you help me determine how many orders, placed in the quarter starting June 1, 1997, had at least one lineitem delivered after its committed date? I need the results grouped by order priority, with the count of such orders sorted in ascending order by order priority.

```sql
FROM
  orders
|> WHERE
     o_orderdate >= date '1997-06-01'
     AND o_orderdate < date_add(date '1997-06-01', INTERVAL 3 month)
     AND EXISTS(
       FROM lineitem
       |> WHERE
            l_orderkey = o_orderkey
            AND l_commitdate < l_receiptdate)
|> AGGREGATE COUNT(*) AS order_count
   GROUP AND ORDER BY o_orderpriority;
```

### 4. Potential Part Promotion Query
Which suppliers in Peru supply parts whose names begin with 'tan' and have an excess inventory of these parts-where excess is defined as having available quantity greater than 50% of the total quantity shipped in 1996? Please return the supplier's name and address, sorted in alphabetical order by name.

```sql
FROM
  supplier,
  nation
|> WHERE
     s_suppkey IN (
       FROM
         partsupp,
         part
       |> WHERE p_name LIKE 'tan%'
       |> WHERE
            ps_partkey = p_partkey
            AND ps_availqty > (
              FROM lineitem
              |> WHERE
                   l_partkey = ps_partkey
                   AND l_suppkey = ps_suppkey
                   AND l_shipdate >= date '1996-01-01'
                   AND l_shipdate < date_add(date '1996-01-01', INTERVAL 1 year)
              |> AGGREGATE 0.5 * sum(l_quantity))
       |> SELECT ps_suppkey)
     AND s_nationkey = n_nationkey
     AND n_name = 'PERU'
|> SELECT s_name, s_address
|> ORDER BY s_name;
```

### 5. Customer Distribution Query 
Can you generate a report that shows the distribution of customers based on the number of orders they have placed? Include customers with zero orders, and make sure to exclude any orders where the comment contains the text 'unusual packages'. The output should list, for each order count, how many customers have that many orders, sorted by the number of customers (in descending order) and then by the order count (also in descending order).

```sql
FROM customer
|> LEFT OUTER JOIN orders ON c_custkey = o_custkey AND o_comment NOT LIKE '%unusual%packages%'
|> AGGREGATE COUNT(o_orderkey) c_count
   GROUP BY c_custkey
|> AGGREGATE COUNT(*) AS custdist
   GROUP BY c_count
|> ORDER BY
     custdist DESC,
     c_count DESC;
```

### 6. Discounted Revenue Query
Can you calculate the gross discounted revenue for orders where the parts meet any of the following criteria?
- Parts of brand 'Brand#53' contained in 'SM CASE', 'SM BOX', 'SM PACK', or 'SM PKG', with a quantity between 5 and 15 and a size between 1 and 5.
- Parts of brand 'Brand#41' contained in 'MED BAG', 'MED BOX', 'MED PKG', or 'MED PACK', with a quantity between 15 and 25 and a size between 1 and 10.
- Parts of brand 'Brand#21' contained in 'LG CASE', 'LG BOX', 'LG PACK', or 'LG PKG', with a quantity between 29 and 39 and a size between 1 and 15.
Additionally, only consider orders that were shipped by air (i.e., with a shipping mode of 'AIR' or 'AIR REG') and were delivered in person. The revenue should be computed as the sum of l_extendedprice * (1 - l_discount) for all orders that qualify.

```sql
FROM
  lineitem,
  part
|> WHERE
     # Added this because optimizer is needed to pull this out of the OR.
     p_partkey = l_partkey
     AND (
       (
         p_partkey = l_partkey
         AND p_brand = 'Brand#53'
         and p_container in ('SM CASE', 'SM BOX', 'SM PACK', 'SM PKG')
         AND l_quantity >= 5
         AND l_quantity <= 5 + 10
         AND p_size BETWEEN 1 AND 5
        and l_shipmode in ('AIR', 'AIR REG')
         AND l_shipinstruct = 'DELIVER IN PERSON')
       OR (
         p_partkey = l_partkey
         AND p_brand = 'Brand#41'
         and p_container in ('MED BAG', 'MED BOX', 'MED PKG', 'MED PACK')
         AND l_quantity >= 15
         AND l_quantity <= 15 + 10
         AND p_size BETWEEN 1 AND 10
         and l_shipmode in ('AIR', 'AIR REG')
         AND l_shipinstruct = 'DELIVER IN PERSON')
       OR (
         p_partkey = l_partkey
         AND p_brand = 'Brand#21'
         and p_container in ('LG CASE', 'LG BOX', 'LG PACK', 'LG PKG')
         AND l_quantity >= 29
         AND l_quantity <= 29 + 10
         AND p_size BETWEEN 1 AND 15
         and l_shipmode in ('AIR', 'AIR REG')
         AND l_shipinstruct = 'DELIVER IN PERSON'))
|> AGGREGATE
     sum(l_extendedprice * (1 - l_discount)) AS revenue;
```

### 7. Global Sales Opportunity Query
Can you identify the potential global sales opportunities by finding customers from specific regions-where the country code is defined as the first two digits of their phone number (i.e., one of '10', '19', '14', '22', '23', '31', '13') - who have not placed any orders and whose account balance is greater than the average positive account balance for these regions? For each country code, please return the number of such customers and the total account balance, sorting the results by the country code.

```sql
FROM customer
|> WHERE
     substr(c_phone, 1, 2) IN ('10', '19', '14', '22', '23', '31', '13')
     AND c_acctbal > (
       SELECT avg(c_acctbal)
       FROM customer
       WHERE
         c_acctbal > 0.00
         AND substr(c_phone, 1, 2) IN ('10', '19', '14', '22', '23', '31', '13')
     )
     AND NOT EXISTS(
       FROM orders
       |> WHERE o_custkey = c_custkey
     )
|> AGGREGATE
     COUNT(*) AS numcust,
     sum(c_acctbal) AS totacctbal
   GROUP AND ORDER BY substr(c_phone, 1, 2) AS cntrycode;
```

### 8. Suppliers Who Kept Orders Waiting Query
Which suppliers in Peru were solely responsible for delaying shipments in multi-supplier orders with a final status of 'F'? For each supplier, count the number of orders where they failed to meet the committed delivery date-while every other supplier on the same order delivered on time. Please list the supplier names along with the count of such delayed orders, ordered from the highest number of delays to the lowest, and show only the top 100 suppliers.

```sql
FROM
  supplier,
  lineitem l1,
  orders,
  nation
|> WHERE
     s_suppkey = l1.l_suppkey
     AND o_orderkey = l1.l_orderkey
     AND o_orderstatus = 'F'
     AND l1.l_receiptdate > l1.l_commitdate
     AND EXISTS(
       FROM lineitem l2
       |> WHERE
            l2.l_orderkey = l1.l_orderkey
            AND l2.l_suppkey <> l1.l_suppkey)
     AND NOT EXISTS(
       FROM lineitem l3
       |> WHERE
            l3.l_orderkey = l1.l_orderkey
            AND l3.l_suppkey <> l1.l_suppkey
            AND l3.l_receiptdate > l3.l_commitdate)
     AND s_nationkey = n_nationkey
     AND n_name = 'PERU'
|> AGGREGATE COUNT(*) AS numwait
   GROUP BY s_name
|> ORDER BY
     numwait DESC,
     s_name
|> LIMIT 100;
```

# Task
Your task is to generate SQL in BigQuery's pipe query syntax described above based on questions in natural language and the presented database structure.

You will then write out your thought process in detail followed by a single SQL query enclosed in ```sql ...``` that answers the question.

BigQuery database schema:
| song : singer_id [number] (1, 2, 4), title [text] ('Do They Know It's Christmas', 'F**k It (I Don't Want You Back)', 'Cha Cha Slide'), song_id [number] (1, 2, 3), sales [float] (1094000.0, 552407.0, 300000.0), highest_position [float] (1.0, 3.0) | singer : birth_year [float] (1944.0, 1948.0, 1949.0), citizenship [text] ('France', 'United States', 'Chile'), name [text] ('Liliane Bettencourt', 'Christy Walton', 'Alice Walton'), singer_id [number] (1, 2, 3), net_worth_millions [float] (30.0, 28.8, 26.3) |

The question: List the name of singers in ascending order of net worth.
\end{lstlisting}

\subsection{Error Classification}\label{appendix:error_prompt}
The prompt used for Section~\ref{sec:analysis} simply outlines the taxonomy described in Appendix~\ref{appendix:taxonomy}.

\lstset{
  backgroundcolor=\color{backcolour2},
}

\begin{lstlisting}
You are provided with the following information:

- **Natural Language Question:** A question posed by a user.
- **Predicted SQL Query:** SQL query generated by a text-to-SQL model.
- **Execution Results of Predicted Query:** The results of executing the predicted query (or an error message).
- **Gold Standard SQL Query:** The correct SQL query.
- **Gold Standard Query Execution Results:** Execution results of the correct (gold standard) SQL query.

Analyze the predicted SQL query and determine the type of error based on provided execution results and query structures. Classify the error into one of the following categories:

- **Dialect:** Issues due to differences between SQL dialects.
- **Schema Linking:** Incorrect matching of natural language terms to schema elements (e.g., hallucinated or incorrect columns or tables).
- **Data Type:** Issues arising from mismatches or unexpected data types within the database.
- **Aggregation:** Incorrect use or omission of aggregation functions (e.g., COUNT, SUM), leading to inaccurate summarization.
- **Logical Form and Condition:** Incorrect logical query structures, missing conditions, inappropriate filters, or incorrect ordering logic.
- **Table Joins:** Incorrect or missing table joins, irrelevant tables, or misidentified join conditions.
- **Projection:** Overall correct queries with incorrect columns selected.

- **Other:** Does not fit into any of the categories above.

Provide your analysis in the following format:

```json
{
  "error_category": "Projection" | "Dialect" | "Schema Linking" | "Logical Form and Condition" | "Data Type" | "Aggregation" | "Table Joins" | "Other",
  "reasoning": "Brief explanation supporting your classification."
}
```

\end{lstlisting}
\end{document}